# Neural Network Implementation of the Renormalization Group for Fault Diagnosis with Class Imbalance

Evgeny Nikulchev, Dmitry Ilin
MIREA – Russian Technological University, Moscow, Russia
e-mail: nikulchev@mail.ru

***Abstract*— The application of machine learning models in practical tasks faces challenges such as class imbalance and multidimensional noise. This paper proposes RGNet, a neural network architecture based on the concept of the renormalization group (RG), for hierarchical coarse-graining of the feature space. The model sequentially compresses the input dimensionality and concatenates all scales before classification, allowing it to capture both local details and global patterns. The notion of RG-flows is introduced — interpretable low-dimensional representations whose visualization via t-SNE reveals a discrete curvilinear structure confirming the effectiveness of coarse-graining. Experimental results are presented on the imbalanced AI4I dataset. The obtained results demonstrate that RGNet is a universal, interpretable, and competitive solution for fault prediction in applications with imbalanced classes.**

## I. INTRODUCTION

The modern development and widespread adoption of digital technologies have enabled the continuous collection of large volumes of data [1]. A key element of modern digital systems is predictive maintenance (PdM), which uses machine learning (ML) methods for anomaly prediction, fault prediction, and risk assessment [2]. Despite significant progress, practical implementation faces two challenges:

1) Strong class imbalance (in real industrial datasets, examples of failures (risks) often constitute less than 5% of the total sample [3]). This leads standard ML algorithms, which optimize overall accuracy, to tend to ignore the rare fault class by predicting normal state for all objects [4]. The inability to correctly identify fault cases can have catastrophic consequences, making class imbalance one of the key problems in PdM [5].

2) Multiscale nature and noise of data. Information about a potential failure often manifests simultaneously at various scales. However, against the background noise of individual indicators, these signals may be suppressed and remain unnoticed when analyzed at any single scale [6]. Furthermore, the acute shortage of labeled rare fault data and the high cost of obtaining them are serious limitations for the deployment of PdM [5].

Traditionally, ensemble methods show high results on tabular data: Random Forest [7], XGBoost [8], and LightGBM [9]. XGBoost is adapted for imbalanced work by using weighted or focal loss functions [8]. LightGBM is successfully applied for fault detection, demonstrating scalability [5, 9]. Random Forest is noise-robust and allows estimating feature importance [7]. Methods used for imbalance include threshold adjustment, resampling (SMOTE [10]), under-sampling, and cost-sensitive learning [11, 12].

Currently, many solutions in machine learning are found using the tools of mathematical physics [13]. This paper proposes using the renormalization group (RG), widely applied in quantum field theory, as a conceptual theoretical foundation. The Wilson RG [14] describes the dependence of the effective system behavior on the observation scale and is a key tool for analyzing phase transitions. Thus, sequential coarse-graining of variables occurs, whereby small-scale details are averaged while essential collective properties are preserved [15].

There exist works at the intersection of RG and ML. Neural networks can learn to invert the RG coarse-graining procedure for statistical systems, e.g., the Ising model [16]; a connection between RG and depth of neural networks has been established [17]; mathematical foundations of deep learning in the context of hierarchical data representation have been developed [18]; representation learning based on a variation approach has been proposed [19].

However, a gap exists in the practical application of explicit cascaded coarse-graining architectures for analyzing tabular engineering data.

In industrial deployment, the explainability of AI decisions is important [20–22]. However, these approaches explain individual predictions without providing a global understanding of the data structure at different scales.

The aim of the paper is to develop theoretical foundations for building RGNet for ML tasks with strongly expressed class imbalance that are important for practical applications.

The contributions of this paper are as follows:

- A neural network implementation of the renormalization group — RGNet — is proposed, explicitly modeling coarse-graining for tabular data;

- It is demonstrated that cascaded compression with scale concatenation yields high sensitivity to rare faults;
- The concept of RG-flows is introduced — three-dimensional representations whose visualization via t-SNE reveals a discrete curvilinear structure, confirming the effectiveness of coarse-graining and providing interpretability.

The paper is organized as follows. Section II describes the methodology and architecture of RGNet; Section III presents results on the AI4I2020 dataset; Section IV is devoted to analysis of RG-flows; Section V discusses advantages and limitations; Section VI concludes the work.

## II. METHODOLOGY AND ARCHITECTURE OF RGNET

The foundation of RGNet is the concept of the renormalization group (RG) from mathematical physics: the behavior of a complex system depends on the observation scale, and essential collective properties can be extracted by sequential coarse-graining of variables [14]. Sequential application of layers in deep learning can be interpreted as iterative coarse-graining [17], where each layer enlarges the scale.

The proposed RGNet architecture is based on two key RG principles.

**1.** Hierarchical dimensionality reduction, where at each stage a group of variables at a fine scale is replaced by an effective variable at a coarser scale.

**2.** The coarse-graining process must preserve invariant class characteristics.

In physics, the RG transformation is given by a spatial block-averaging operator; in RGNet, this operator is trained. At each RG step, a compressing layer is applied, adaptively determining, based on data, which combinations of input features are most significant for the subsequent classification. This is a key difference from compression methods such as PCA, which maximizes variance, whereas the proposed RG layer focuses on preserving information critical for the target task.

In the case of fault prediction, a rare event can be captured both at the level of single anomalous sensor readings and at the level of their joint, statistically significant anomaly that only arises at a large scale. Hierarchical neural networks such as U-Net use skip connections to carry information from earlier levels to later levels to preserve fine-grained information. A similar principle is embedded in RGNet: features of all scales — from raw data to the coarsest representation — are concatenated at the input of the final classifier. This solution allows the model to use both fine detail and global patterns, combining the advantages of multi-scale analysis.

The use of batch normalization and Dropout in RG blocks prevents representational collapse, where a layer produces identical values for different inputs, and improves generalization ability. Thus, invariance, regularization, and stability are achieved.

Let us formulate the task formally.

Let the initial feature vector be denoted $h^{(0)} = x \in \mathbb{R}^{d_0}$. Define an RG layer as a dimensionality-reducing transformation:

$$h^{(k+1)} = \phi\left(W^{(k)} h^{(k)} + b^{(k)}\right), \qquad k = 0,1,\dots,L-1,$$

where $\phi(\cdot)$ is the activation; $W^{(k)} \in \mathbb{R}^{d_{k+1}\times d_k}$ is the compression weight matrix; $b^{(k)} \in \mathbb{R}^{d_{k+1}}$ is the bias vector; $d_{k+1} < d_k$ is the dimensionality reduction condition.

Unlike a traditional hidden layer, the dimensions $d_k$ are explicitly set as a decreasing sequence.

Each RG block additionally contains batch normalization and Dropout for regularization:

$$\tilde{h} = \mathrm{Dropout}\left(\mathrm{BN}(W_1 h + b_1)\right), \quad h_{\mathrm{out}} = \mathrm{BN}\left(W_2 \tilde{h} + b_2\right),$$

where BN is batch normalization, Dropout coefficient is 0.1. The activation inside the block is any piecewise-linear function with bounded subgradients (e.g., ReLU, Leaky ReLU); the output of the block is linear.

After all RG layers, concatenation of all scales is performed:

$$f = \left[h^{(0)}; h^{(1)}; \dots; h^{(L)}\right] \in \mathbb{R}^D, \quad D = \sum_{k=0}^{L} d_k.$$

The classifier is a two-layer neural network:

$$z_1 = \mathrm{Dropout}\left(\mathrm{BN}(W_{c1} f + b_{c1})\right),$$
$$z_2 = \mathrm{BN}(W_{c2} z_1 + b_{c2}),$$
$$\hat{y} = \sigma(w_{\mathrm{out}}^{\top} z_2 + b_{\mathrm{out}}),$$

where $\sigma(z) = 1/(1+e^{-z})$ is the sigmoid function.

For any initial x, we define the trajectory $h^{(1)}(x), h^{(2)}(x)$.

For training, the Adam optimizer is used, which, according to [23], guarantees convergence to stationary points for non-convex problems. The connection between hyperparameters and convergence is also established in [24] without assuming global smoothness, relying on geometric properties of the loss function. Reference [25] shows two-phase convergence: sublinear, then exponential.

**Lemma 1** (Scale invariance and stability of RGNet).

Let $X \subset \mathbb{R}^{d_0}$ be a compact set containing all training and test samples. For each scale $k = 0,1,\dots,L-1$ define the RG transformation $F_k: \mathbb{R}^{d_k} \to \mathbb{R}^{d_{k+1}}$,

$$F_k(h) = \mathrm{BN}\left(W_k^{(2)} \cdot \mathrm{Dropout}\left(\mathrm{BN}(W_k^{(1)} h + b_k^{(1)})\right)\right) + b_k^{(2)},$$

where BN is batch normalization, Dropout rate is 0.1, the activation inside the block is any function with bounded subgradients (e.g., ReLU), and the block output is linear. The total composition $F = F_{L-1} \circ \dots \circ F_0$ maps $X$ into $\mathbb{R}^{d_L}$.

Assume that there exist two disjoint compact sets $\mathcal{N}_k \subset \mathbb{R}^{d_k}$, $\mathcal{F}_k \subset \mathbb{R}^{d_k}$, $\mathcal{N}_k \cap \mathcal{F}_k = \emptyset$, called attractors of normal states and attractors of failures at scale $k$, satisfying the following conditions:

Initial attractors – images of original data:

$$\mathcal{N}_0 = \{x \in X : y = 0\}, \qquad \mathcal{F}_0 = \{x \in X : y = 1\}.$$

Invariance under coarse-graining – for each layer $k = 0, \ldots, L-1$ we have

$$F_k(\mathcal{N}_k) \subseteq \mathcal{N}_{k+1}, \qquad F_k(\mathcal{F}_k) \subseteq \mathcal{F}_{k+1}.$$

(Class membership is preserved when moving to a coarser scale.)

Separation at the coarsest scale:

$$\rho = \inf\{\| u - v \| : u \in \mathcal{N}_L,\ v \in \mathcal{F}_L\} > 0.$$

Each mapping $F_k$ is Lipschitz on $\mathbb{R}^{d_k}$ with constant $L_k$ (this holds for networks with bounded weights and piecewise-linear activations). Denote the overall Lipschitz constant of the composition as

$$L_{\max} = \prod_{k=0}^{L-1} L_k < \infty.$$

Then the following statements hold:

Noise robustness (class stability). Let $x \in \mathcal{N}_0$ be a normal sample, $\tilde{x} = x + \epsilon$, $\epsilon \sim \mathcal{N}(0, \sigma^2 I_{d_0})$. For any $\delta \in (0,1)$ with probability at least $1 - \delta$ we have

$$\mathrm{dist}(F(\tilde{x}), \mathcal{N}_L) \le L_{\max}\, \sigma\big(\sqrt{d_0} + \sqrt{2\ln(1/\delta)}\big).$$

If additionally

$$L_{\max}\, \sigma\big(\sqrt{d_0} + \sqrt{2\ln(1/\delta)}\big) < \frac{\rho}{2},$$

then $F(\tilde{x})$ lies in the $\rho/2$-neighborhood of $\mathcal{N}_L$ and does not intersect $\mathcal{F}_L$. An analogous statement holds for failures $x \in \mathcal{F}_0$.

Preservation of class separability with depth. For any $x_{\text{norm}} \in \mathcal{N}_0$ we have $F(x_{\text{norm}}) \in \mathcal{N}_L$; for any $x_{\text{fail}} \in \mathcal{F}_0$ we have $F(x_{\text{fail}}) \in \mathcal{F}_L$. Hence, even after strong compression (small $d_L$) the classes remain distinguishable.

**Proof**.

Noise robustness. Since $F$ is Lipschitz with constant $L_{\max}$, for any $x, \tilde{x}$ we have

$$\| F(\tilde{x}) - F(x) \| \le L_{\max} \| \tilde{x} - x \| = L_{\max} \| \epsilon \|.$$

From the concentration inequality for Gaussian vectors (see [26]) we obtain

$$\mathbb{P}\big(\| \epsilon \| \ge \sigma(\sqrt{d_0} + t)\big) \le e^{-t^2/2}.$$

Choose $t = \sqrt{2\ln(1/\delta)}$. Then with probability at least $1 - \delta$,

$$\| \epsilon \| \le \sigma\big(\sqrt{d_0} + \sqrt{2\ln(1/\delta)}\big).$$

Hence,

$$\mathrm{dist}(F(\tilde{x}), \mathcal{N}_L) \le \| F(\tilde{x}) - F(x) \| + \mathrm{dist}(F(x), \mathcal{N}_L).$$

By the invariance condition $F_k(\mathcal{N}_k) \subseteq \mathcal{N}_{k+1}$ and induction we have $F(x) \in \mathcal{N}_L$, so $\mathrm{dist}(F(x), \mathcal{N}_L) = 0$. The first inequality is proved.

If in addition $L_{\max} \| \epsilon \| < \rho/2$, then $\| F(\tilde{x}) - F(x) \| < \rho/2$. Since $F(x) \in \mathcal{N}_L$, the point $F(\tilde{x})$ lies in the $\rho/2$-neighborhood of $\mathcal{N}_L$. The distance between $\mathcal{N}_L$ and $\mathcal{F}_L$ is $\rho$, therefore this neighborhood contains no points from $\mathcal{F}_L$. Thus classification is stable under small noise. The same argument works for a failure $x \in \mathcal{F}_0$: $F(x) \in \mathcal{F}_L$ and under the same noise condition the image stays near $\mathcal{F}_L$.

Preservation of class separability. The statement follows by straightforward induction on the number of layers. Base $k = 0$ holds by definition of $\mathcal{N}_0, \mathcal{F}_0$. Inductive step: if $x \in \mathcal{N}_k$ then $F_k(x) \in \mathcal{N}_{k+1}$ by condition 2. The same for failures. Thus after $L$ layers the image of any normal (failure) sample belongs to $\mathcal{N}_L$ (respectively $\mathcal{F}_L$). Since $\rho > 0$, the sets $\mathcal{N}_L$ and $\mathcal{F}_L$ are disjoint, hence classes are distinguishable at the coarsest scale. □

**Remark 1** (Gradient flow in RGNet). Although not required for the formal statements above, the concatenation $f = [h^{(0)}; h^{(1)}; \ldots; h^{(L)}]$ at the classifier input provides favorable gradient properties. Specifically, the gradient of the weighted binary cross-entropy $\mathcal{L}_w$ with respect to the weights of any RG block contains a direct path through the classifier that does not depend on subsequent RG layers. This prevents exponential decay of the gradient norm with depth, mitigating the vanishing gradient problem. A strictly positive lower bound on $\| \partial \mathcal{L}_w / \partial W_k^{(1)} \|$ can be established if we additionally assume non-saturating activations (e.g., Leaky ReLU with $\alpha > 0$). In our experiments with ReLU we observe stable convergence (Fig. 1), indicating that in practice the gradient flow remains sufficient. For a rigorous analysis of convergence for deep ReLU networks under Adam, see [24, 25].

**Remark 2**. The invariance conditions $\mathrm{F_k}(\mathcal{N}_\mathrm{k}) \subseteq \mathcal{N}_{\mathrm{k+1}}$ and $\mathrm{F_k}(\mathcal{F}_\mathrm{k}) \subseteq \mathcal{F}_{\mathrm{k+1}}$ do not assume contraction of each layer. They only require that coarse-graining does not mix the classes. This is a substantially weaker and more realistic assumption, which can be verified experimentally in the original paper, where visualization of RG-flows confirms the existence of separate clusters for failures). Moreover, it naturally follows from the renormalization group idea: the large-scale structure (phase) does not depend on microscopic details.

Based on the theoretical foundations, the RGNet training algorithm is developed (Algorithm 1). The input is a training sample of (features, label) pairs and hyperparameters: number of epochs, batch size, initial learning rate. First, weights of all layers (RG blocks and classifier) are initialized. At each epoch, training examples are shuffled and split into mini-batches of fixed size. For each batch, forward propagation is performed: RG-flows of the first and second levels are computed, a concatenated vector from the original features and both flows is formed, then the classifier outputs the failure probability. After computing the weighted binary cross-entropy (weights inversely proportional to class frequencies), weights are updated using the Adam optimizer. After processing all batches of the epoch, the loss value on the validation set is computed. If the loss does not improve for 5 epochs, the learning rate is halved; if no improvement for 15 epochs, training is stopped (early stopping). The output is the trained RGNet model.

**Algorithm 1:** Training of RGNet

**Input**: training sample $(x_i, y_i)$, $i = 1..N$; hyperparameters (number of epochs, batch size $B$, learning rate $lr$)

Initialize weights of RG blocks and classifier
Shuffle training examples
Split into batches of size $B$
For{ each epoch }{
  For{each epoch каждого батча ($X_b$, $y_b$)}{
    Compute $h^{(1)} = R_1(X_b), h^{(2)} = R_2(h^{(1)})$
    Form $f = [X_b; h^{(1)}; h^{(2)}]$
    Obtain predictions $\hat{y}$
    Compute weighted binary cross-entropy loss $L$
    Update weights using Adam
  End
  Evaluate $L$ on validation set
  If $L$ does not improve for 5 epochs, halve $lr$
  If $L$ does not improve for 15 epochs, stop training
**End**

**Output**: trained RGNet model

Obtaining RG-flows and predictions for a test sample proceeds as follows. A test sample (feature vector) and the trained RGNet are fed as input. The sample passes through RG blocks, resulting in RG-flows. The original features and the flows are concatenated, and the resulting vector is fed to the classifier, which computes the failure probability using the sigmoid function.

Consider a verification example of RGNet without class imbalance. The aim is to show that the results are comparable to popular ML methods.

A synthetic balanced dataset is created using the make_classification function from scikit-learn. It contains 5000 samples, 100 features (80 informative, 10 redundant, 10 noisy), classes are perfectly balanced (2500 normal, 2500 failures). Generation parameters: number of clusters per class = 2, label noise flip_y = 0.01, class separability class_sep = 1.0. Split into train/test 80/20.

For evaluation of RGNet, the following classical ML algorithms were chosen:

- XGBoost with parameter scale_pos_weight = 28.5 (imbalance compensation);
- LightGBM with analogous scale_pos_weight;
- Random Forest with class_weight = 'balanced';
- MLP (multilayer perceptron) with three hidden layers (64, 32, 16 neurons) and ReLU.

All baseline models were trained on the same data as RGNet.

Each model was evaluated on the test set using the following metrics:

- Recall for failure class = TP / (TP + FN);
- Precision for failure class = TP / (TP + FP);
- F1-score = 2 · Precision · Recall/(Precision + Recall);
- Accuracy = (TP + TN) / (TP + TN + FP + FN).

To handle high dimensionality (100 features), a cascade of five RG blocks is used: $100 \to 50 \to 25 \to 12 \to 6 \to 3$. Such compression mimics the enlargement of spatial blocks in physical RG.

Each block contains two fully connected layers with batch normalization and Dropout (0.1). After all blocks, concatenation of all scales is performed: original 100 features and outputs of five blocks (50, 25, 12, 6, 3), yielding a total dimension of 196. The classifier consists of two layers (64 and 32 neurons) with Dropout 0.3 and output sigmoid. Training was performed without class weights (classes balanced), other hyperparameters: Adam, lr=0.001, batch 64, early stopping. Training is stable, loss decreases, accuracy reaches 0.98 by the 20th epoch, early stopping prevents overfitting.

Table I reports the results of RGNet and baseline methods.

TABLE I. COMPARISON OF METHODS ON SYNTHETIC DATASET (100 FEATURES)

| Method | Accuracy | Recall | Precision | F1 |
|---|---|---|---|---|
| **RGNet** | 0.98 | 0.97 | 0.98 | 0.98 |
| XGBoost | 0.92 | 0.91 | 0.92 | 0.92 |
| LightGBM | 0.90 | 0.90 | 0.90 | 0.90 |
| Random Forest | 0.89 | 0.89 | 0.89 | 0.89 |
| MLP | 0.96 | 0.96 | 0.95 | 0.96 |

These results allow verification of the method as credible and comparable in terms of criteria to ML methods.

## III. EXPERIMENTAL STUDY

Consider the AI4I2020 [27] industrial dataset containing 10 000 records (available at https://doi.org/10.24432/C5HS5C). Each record is described by five physical features: air temperature, process temperature, rotational speed, torque, tool wear. The target is binary: 0 (normal) or 1 (failure). Class distribution is strongly imbalanced: normal – 9661 (96.6%), failure – 339 (3.4%). Data were split into training (80%) and test (20%) stratified by label.

An RG architecture $5 \to 4 \to 3$ was developed. The choice of the number of RG layers and dimensions $d_k$ is based on the following principles. Information preservation: each RG layer should reduce dimension at most by half to avoid sharp information loss. For $d_0 = 5$, a natural first step is $d_1 = 4$ (reduction by 1), then $d_2 = 3$.

Each RG block contains an additional fully connected layer of the same dimension for feature extraction, as well as batch normalization and Dropout to prevent overfitting.

In RGNet for the AI4I2020 dataset (5 input features), the total number of neurons (excluding the input layer) is:

• RG Block 1: two fully connected layers of 4 neurons each → 8;

• RG Block 2: two fully connected layers of 3 neurons each → 6;

• Classifier: two hidden layers (64 and 32 neurons) → 96.

• Output layer: 1 neuron (sigmoid).

Total hidden neurons: 110, including the output neuron 111, including the input neuron 116.

In our experiments, we use ReLU activations inside RG blocks. Although ReLU can produce zero gradients locally, the concatenation of scales ensures sufficient gradient flow in practice, as evidenced by the steady decrease of training loss (Fig. 1).

Training uses the Adam optimizer (initial learning rate 0.001), ReduceLROnPlateau schedule (factor 0.5, patience=5), batch size 64. Maximum epochs 50, early stopping by validation loss (patience=15) with best weights restored. Additional regularization: L2 penalty ($10^{-4}$) on all dense layers. For AI4I2020, $w_1 \approx 14.5$, $w_0 = 0.5$.

For AI4I2020, threshold classifiers based on feature coincidence were implemented: a sample is considered a failure if at least $m$ out of 5 features ($m = 2{,}3$) fall into the interval [28] $[\text{mean}_j - k\sigma_j,\ \text{mean}_j + k\sigma_j]$, where mean and $\sigma$ are computed from training failures, and $k$ is tuned on validation. A calibrated RGNet with threshold 0.92 was also considered.

The comparison results with methods using AI4I2020 are presented in Table II.

TABLE II. COMPARISON OF METHODS ON AI4I2020TEST SET

| Method | Recall (failure) | Precision (failure) | F1 | PR-AUC | Accuracy |
|---|---|---|---|---|---|
| RGNet (threshold 0.5) | 0.9265 | 0.2582 | 0.404 | 0.670 | 0.907 |
| XGBoost | 0.750 | 0.680 | 0.713 | 0.802 | 0.932 |
| LightGBM | 0.764 | 0.642 | 0.698 | 0.808 | 0.956 |
| Random Forest | 0.558 | 0.884 | 0.685 | 0.791 | 0.902 |
| MLP | 0.588 | 0.800 | 0.678 | 0.785 | 0.915 |
| RGNet (threshold 0.92) | 0.5882 | 0.6780 | 0.630 | 0.670 | 0.977 |
| Threshold 2/5 (k=0.75) | 0.941 | 0.036 | 0.069 | 0.036 | – |
| Threshold 3/5 (k=1.00) | 0.853 | 0.035 | 0.067 | 0.035 | – |

The results are also illustrated in Fig. 1–3 (insert figures). From Fig. 1 it is seen that losses steadily decrease and accuracy increases, with early stopping occurring around epoch 30, preventing overfitting. By tuning the threshold to maximize F1 on validation, the value 0.92 was obtained, which increases precision to 0.678 while lowering recall to 0.588 (F1=0.630) – a balanced option for practical use.

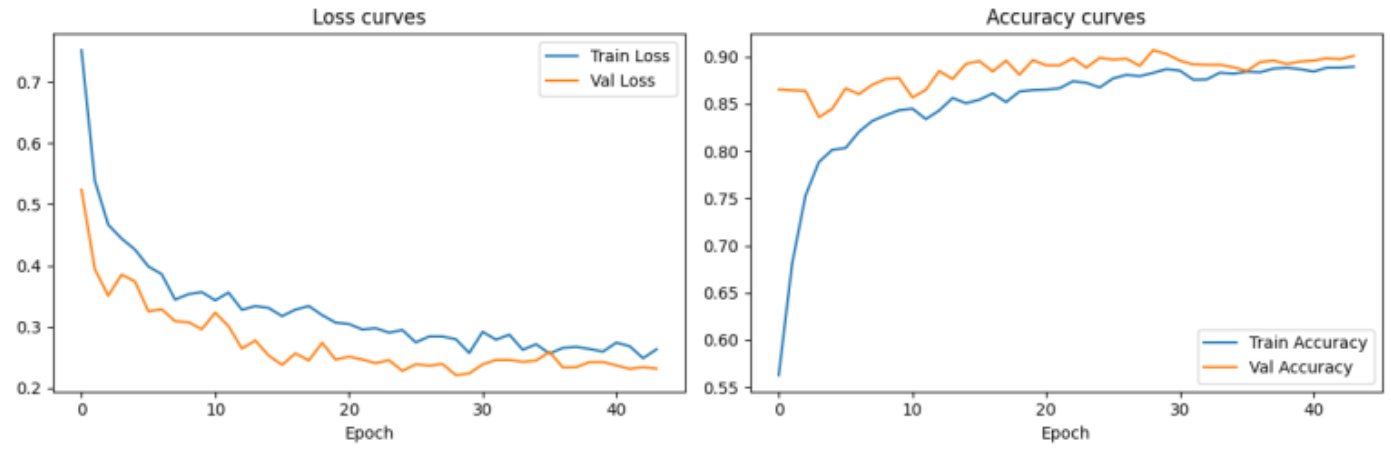

Fig. 1. Loss and accuracy curves RGNet

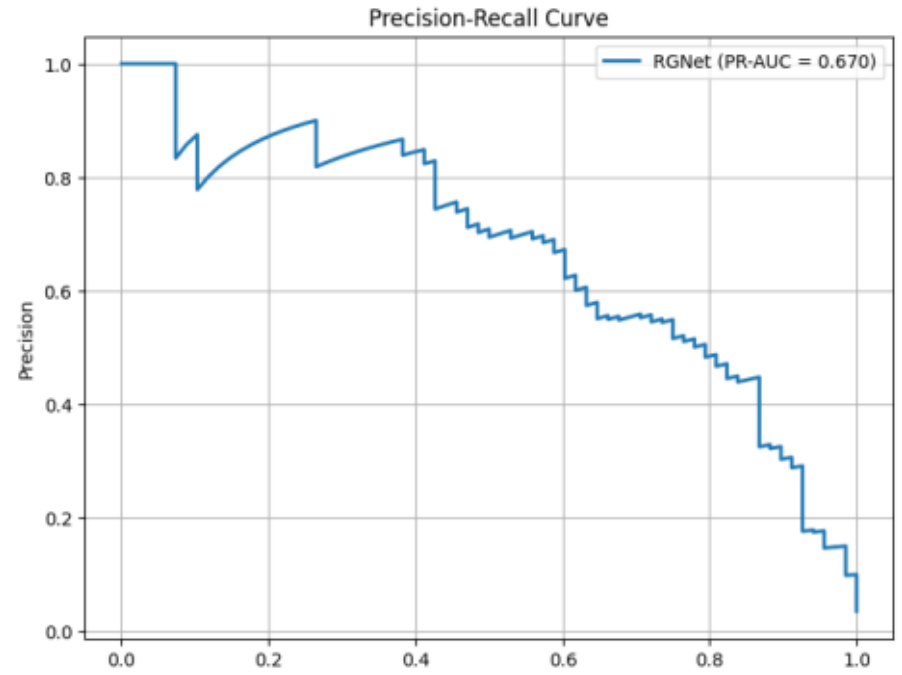

Fig. 2. PR-AUC calculation

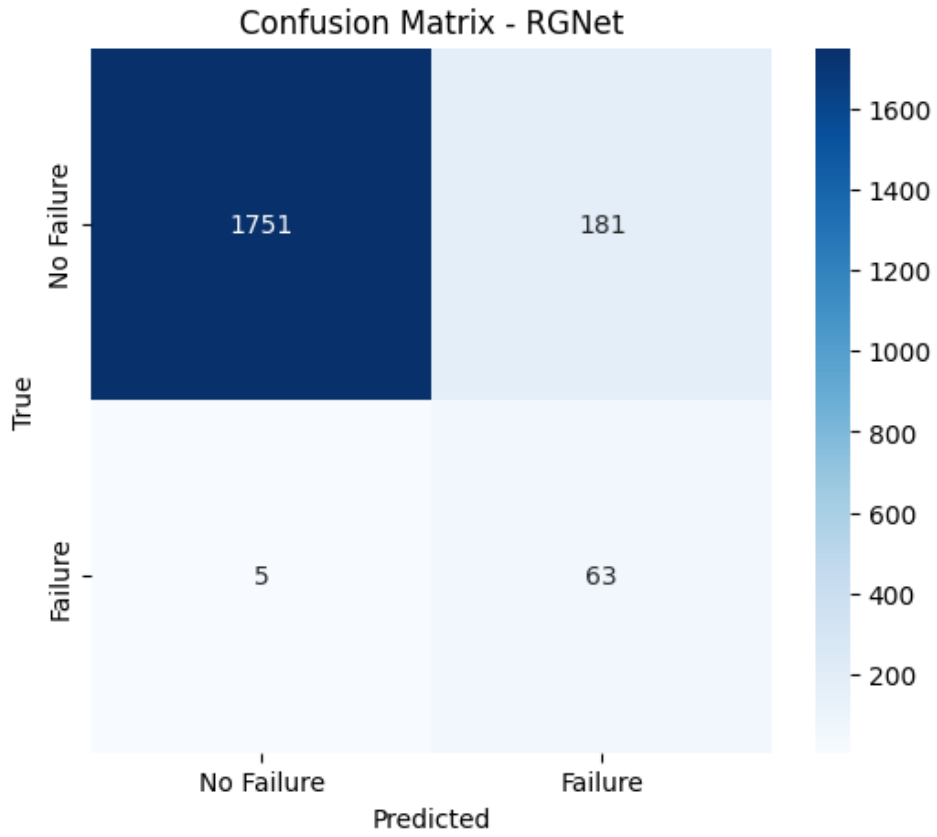

Fig. 3. Confusion matrix

A method for structural analysis of RG-flows is developed.

It is important to note that false alarms in many practical applications are less critical than missing truly “bad” cases. In this case, the model issued a false warning 181 times, which, compared to the data volume, will not significantly hinder the work of technical services responsible for product quality.

An RG-flow of the last scale is defined as the vector $h^{(L)}(x) \in \mathbb{R}^{d_L}$. In the experiment, $d_L = 3$. This is the coarsest representation of the original sample.

For visualization, the three-dimensional RG-flows of all test samples are projected onto a plane using t-SNE [29]. The resulting two-dimensional diagrams reveal a discrete curvilinear structure – a consequence of the limited set of unique RG-flows and t-SNE preserving local distances, connecting close discrete points into continuous lines. Statistical analysis shows that the set of unique vectors is limited (a few tens). Specifically, 187 out of 271 failures in the training set share the same RG-flow $t = (-0.612, -0.553, 0.698)$.

Figure 4 shows the t-SNE projection of three-dimensional RG-flows onto the plane with t-SNE (perplexity = 30, random state = 42). Points are colored by true labels: blue – normal (class 0), red – failure (class 1). Smooth thin curves (the main "highway" for normal) and isolated clusters of failures are visible.

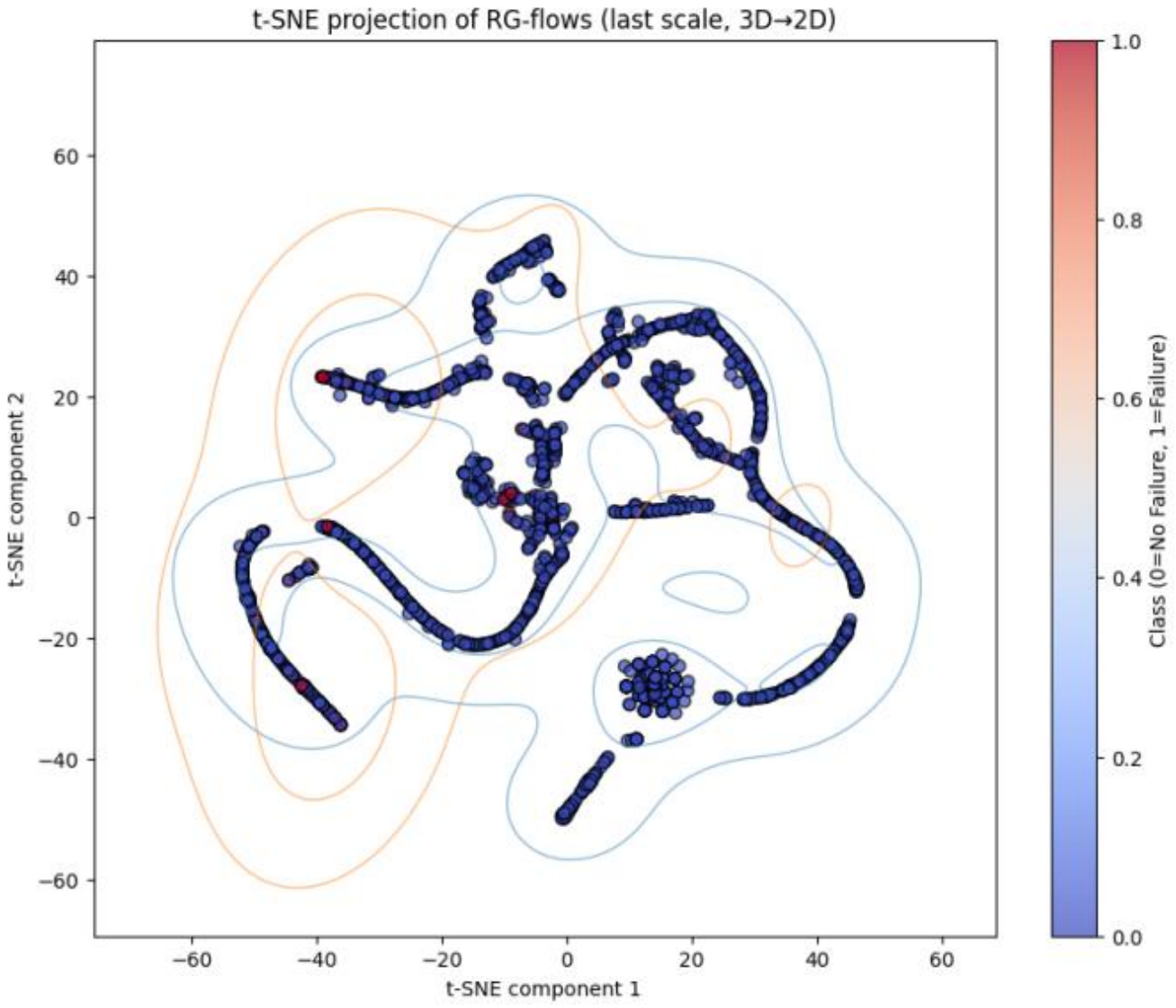

Fig. 4. t-SNE projection of RG-flows

The points do not form a random cloud but align along several smooth curves (resembling parabolas or arcs). This is a consequence of the discrete nature of the RG representation: t-SNE, preserving local distances, connects close discrete points into continuous lines.

The dense cluster of blue points forms the main curve, which corresponds to normal operating modes. The spread along the curve reflects natural variation of parameters (e.g., temperature or rotational speed) within the normal range.

Red points (failures) are distributed in two ways. Some red points lie directly on the main curve interspersed with blue ones. These are failures that at the coarse scale are indistinguishable from normal – their RG-flow coincides with the normal one. Nevertheless, the classifier correctly recognizes these samples using finer scales (original features and output of the first RG layer). The other part of red points forms isolated clusters or separate "outgrowths" on the curves. Such samples have unique RG vectors and are classified by the model with high confidence (failure probability > 0.9). They correspond to failures with pronounced anomalies that manifest already at the coarse scale.

When projecting the original five features or the outputs of a conventional MLP (32 neurons) with t-SNE, a chaotic cloud without any structure is obtained. The appearance of sharp curves is a specific property of the RG representation, confirming the effectiveness of coarse-graining.

## IV. Discussion

Comparison of RGNet with baseline methods. On the AI4I2020 dataset with strong imbalance, RGNet at threshold 0.5 achieves recall 0.9265 – 16–17 percentage points higher than XGBoost (0.750) and LightGBM (0.764). This is achieved through coarse-graining: sequential enlargement of the feature space suppresses high-frequency noise and extracts collective correlations characteristic of failures. Meanwhile, the PR-AUC of RGNet (0.670) is lower than gradient boosting (0.80–0.81), indicating a larger fraction of false positives. However, threshold calibration to 0.92 yields a balanced variant: recall 0.588, precision 0.678, F1 0.630.

RG-flows (three-dimensional representations) are visualized using t-SNE, revealing a discrete curvilinear structure. Isolated clusters of failures correspond to anomalies that the model recognizes with high confidence.

Analysis of RG-flows enables for the application domain:

Tracking how changes in physical parameters (e.g., increase in tool wear or temperature rise) move the point along the curve. If the point enters an isolated cluster region, it signals high failure probability.

Thus, RG-flows serve as a bridge between the "black box" of the neural network and the physical understanding of the process, providing interpretability not available in classical machine learning methods.

## V. Conclusion

This paper proposes RGNet (Renormalization Group Network), a neural network architecture that implements the idea of the renormalization group for hierarchical coarse-graining of the feature space in fault prediction tasks. Both experimental and theoretical justifications of its effectiveness are provided.

*Theoretical results.* A lemma on scale invariance is formulated and proved: if the RG layers preserve class membership under coarse-graining (i.e., the images of normal states and failures remain in disjoint compact attractors at each scale), then the classes remain distinguishable even after strong compression. These statements do not require global contraction of each layer but rely only on a natural property of the renormalization group: the large-scale structure (phase) is invariant under coarse-graining.

*Experimental results.* On a synthetic balanced dataset, RGNet achieves accuracy comparable to classical machine learning methods. On the industrial AI4I2020 dataset with strong class imbalance (3.4% failures), RGNet with threshold 0.5 attains a recall of 0.9265 – 16–17 percentage points higher than gradient boosting. Threshold calibration to 0.92 yields a balanced variant (recall 0.588, precision 0.678, F1 = 0.630), suitable for practical applications where false alarms are less critical than missed failures.

*Interpretability.* The concept of RG-flows is introduced – three-dimensional representations of the original data at the coarsest scale. Visualization via t-SNE reveals a discrete curvilinear structure: normal states align along a main "highway", while failures either merge with it (in which case the classifier relies on finer scales) or form isolated clusters. This visualization provides engineers with the ability to formulate threshold rules (e.g., entering a specific region in the t-SNE projection signals high failure risk) and to track the evolution of equipment state over time – something unavailable to classical black-box models.

*Future research directions*. Automatic selection of the number of RG layers and dimensions, as well as integration of RGNet into real-time systems with online learning.